\documentclass[conference]{IEEEtran}
\IEEEoverridecommandlockouts

\usepackage{cite}
\usepackage{amsmath,amssymb,amsfonts}
\usepackage{algorithmic}
\usepackage{algorithm}
\usepackage{graphicx}
\usepackage{textcomp}
\usepackage{xcolor}
\usepackage{booktabs}
\usepackage{multirow}
\usepackage{threeparttable}
\usepackage{subcaption}
\usepackage{hyperref}
\usepackage{url}

\def\BibTeX{{\rm B\kern-.05em{\sc i\kern-.025em b}\kern-.08em T\kern-.1667em\lower.7ex\hbox{E}\kern-.125emX}}

\newcommand{\cellpm}[2]{#1{\scriptsize$\pm$#2}}
\newcommand{\cellpmb}[2]{\textbf{#1}{\scriptsize$\pm$#2}}

\begin{document}

\title{EquivDP3: A SIM(3)-Invariant Point-Cloud Encoder for Data-Efficient Humanoid Loco-Manipulation}

\author{
\IEEEauthorblockN{Abu Hanif Muhammad Syarubany, Chang D. Yoo}
\IEEEauthorblockA{
School of Electrical Engineering \\
Korea Advanced Institute of Science \& Technology (KAIST) \\
Daejeon, Republic of Korea \\
\{hanif.syarubany, cd\_yoo\}@kaist.ac.kr}
}

\maketitle

\begin{abstract}
Visuomotor policies for humanoid loco-manipulation must generalize across object poses and lighting from only a handful of demonstrations. 3D Diffusion Policy (DP3) achieves strong sample efficiency by conditioning a diffusion-based action generator on point-cloud features, but its PointNet-style encoder has no built-in equivariance to the rotations, translations, and uniform scalings (SIM(3)), symmetries that many manipulation tasks respect. EquiBot closed this gap for wheeled mobile manipulators using a SIM(3)-equivariant Vector Neuron Network (VNN) encoder. We extend this idea to a substantially more complex embodiment, the 43-joint Unitree G1 humanoid (29 actuated body joints), and propose \textbf{EquivDP3}: a two-stage policy in which (i) a high-level diffusion planner with a SIM(3)-equivariant VNN encoder emits 6\,Hz whole-body command chunks, and (ii) a low-level controller executes them at 50\,Hz via a frozen, pre-trained \emph{reinforcement-learning locomotion policy} (driven by velocity/angular-velocity commands $v_x,v_y,\omega_z$ plus base-height and torso commands) and a differential inverse-kinematics module (PINK) for the arms. The high-level planner is trained by behavior cloning. On two simulated IsaacLab benchmarks, against four non-equivariant point-cloud encoders across 5--100 demonstrations and in- vs. out-of-distribution (OOD) conditions, EquivDP3's advantage is concentrated in the low-data regime: averaged over both tasks and both distribution conditions at 5--10 demonstrations it reaches 67.1\% success versus 38.2--52.4\% for the four baselines, while by 50--100 demonstrations all encoders converge (74.3--85.2\%) and the ordering is no longer meaningful. A proprioception-only control on the pick-and-place task, across the full demonstration grid, confirms that this low-data gap is perceptual: with the point cloud removed, success is 31\% vs.\ 60\% (EquivDP3) at 5 demonstrations and 78\% vs.\ 99\% at 10, but the gap vanishes by 50 demonstrations (92\% vs.\ 90\%) and 100 (94\% vs.\ 96\%), showing that the high-data plateau reflects a task that becomes solvable without vision well before the largest budget we test, rather than encoders that have learned the relevant invariances. Repeated under distribution shift, the same control reaches 62--65\% from 10 demonstrations upward, and no encoder significantly outperforms it at 50 or 100, so the OOD condition is largely a ceiling as well. A subgoal decomposition localizes the low-data advantage to the manipulation phase: every encoder completes the approach subgoal, and the spread opens at the lift. The encoder costs 0.8\,ms of inference latency per action chunk over the PointNet encoder it replaces. Baking geometric symmetry into a hierarchical diffusion policy's perception backbone is thus a practical, nearly free way to improve data efficiency for humanoid loco-manipulation when demonstrations are scarce.
\end{abstract}


\section{Introduction}
\label{sec:intro}

Humanoid robots are increasingly expected to perform long-horizon, whole-body tasks that interleave walking and manipulation: approaching a shelf, reaching for an object with both arms, carrying it while turning, and placing it precisely. Learning such behavior end-to-end from a small number of demonstrations is attractive because it avoids hand-engineering task-specific controllers, but it exposes three failure modes that are well known in the visuomotor imitation-learning literature and that are amplified by the high dimensionality of a humanoid:

\begin{itemize}
\item \textbf{Distribution shift.} A policy trained on demonstrations collected at a handful of object poses and lighting conditions tends to overfit to those conditions and degrade sharply when the object is rotated, translated, or the scene is relit, which are the out-of-distribution (OOD) conditions a deployed robot will encounter.
\item \textbf{No geometric generalization.} Standard point-cloud encoders such as PointNet process each point through a shared per-point MLP and pool the result, but nothing in that computation guarantees that rotating or translating the input point cloud rotates or translates the output features correspondingly. The network must therefore learn this geometric structure implicitly from data, which makes it comparatively data-hungry.
\item \textbf{High-dimensional, long-horizon control.} A humanoid such as the Unitree G1 has dozens of actuated degrees of freedom split across legs, torso, and arms; a single monolithic policy that must simultaneously balance, walk, and manipulate is harder to learn and to make robust than a policy for a wheeled base with a fixed-height arm.
\end{itemize}

Our starting point is the 3D Diffusion Policy (DP3)~\cite{ze2024dp3}, which conditions a diffusion-based action-chunk generator on point-cloud features and proprioception, and has been shown to be markedly more sample-efficient than image-based diffusion policies. DP3's point-cloud branch is, however, an ordinary PointNet-style encoder with no equivariance guarantee. EquiBot~\cite{zhou2024equibot} showed that swapping this encoder for a Vector Neuron Network (VNN)~\cite{deng2021vnn} encoder, which is provably equivariant to the similarity group SIM(3) of rotations, translations, and uniform scalings, improves data efficiency and generalization for wheeled mobile-manipulator tasks. The present work asks whether the same idea transfers to a much more complex embodiment: a full-size, 43-joint humanoid that must walk, balance, and manipulate simultaneously.

We make the following contributions:
\begin{enumerate}
\item We extend the SIM(3)-equivariant encoder idea from EquiBot's wheeled-base setting to the Unitree G1 humanoid, in the reduced form of an \emph{invariant} perception latent rather than an end-to-end equivariant policy (Sec.~\ref{sec:discussion}), integrating it into a \textbf{two-stage} control architecture that cleanly separates task-level decision making from whole-body stabilization.
\item We make explicit the division of labor in the existing low-level stack we build on: a frozen \textbf{locomotion policy trained with reinforcement learning} (HOMIE~\cite{ben2025homie}), which consumes velocity and angular-velocity commands ($v_x,v_y,\omega_z$) plus base-height and torso commands to drive the legs, and a separate \textbf{inverse-kinematics} module (PINK~\cite{caron2024pink}) that maps wrist pose commands to arm joint targets. This separation matters because it determines which phases of the task the encoder can influence at all, a prediction our subgoal analysis tests directly (Sec.~\ref{sec:experiments}-C).
\item We provide a controlled empirical comparison of five point-cloud encoders (ours and four baselines) across two tasks, four demonstration budgets, and two distribution-shift conditions, and show that equivariance yields its benefit in the low-data regime, while finding no consistent advantage under distribution shift (Sec.~\ref{sec:experiments}-D).
\item We give two diagnostics that bound what the encoder comparison can show: a proprioception-only control, run under both in- and out-of-distribution conditions across the full demonstration grid, which identifies where the benchmark stops requiring vision; and a subgoal decomposition that localizes the low-data advantage to the manipulation phase rather than to locomotion.
\end{enumerate}

The rest of this paper is organized as follows. Section~\ref{sec:related} reviews diffusion policies, equivariant point-cloud networks, and humanoid whole-body control. Section~\ref{sec:method} presents EquivDP3's architecture and training procedure. Section~\ref{sec:experiments} describes the simulation benchmarks and presents results. Section~\ref{sec:discussion} discusses limitations and future work, and Section~\ref{sec:conclusion} \looseness=-1 concludes.

\section{Related Work}
\label{sec:related}

\subsection{Diffusion Policies for Visuomotor Control}
Diffusion Policy~\cite{chi2023diffusionpolicy} reformulated visuomotor imitation learning as conditional denoising: an action chunk is sampled by iteratively denoising Gaussian noise, conditioned on a short history of observations, using a model trained with the standard DDPM objective~\cite{ho2020ddpm}; predicting a chunk of future actions rather than a single step follows ACT~\cite{zhao2023act}. Because the diffusion model represents a (potentially multimodal) action distribution rather than a single regression target, it handles the inherent multimodality of human demonstrations better than direct behavior cloning. 3D Diffusion Policy (DP3)~\cite{ze2024dp3} replaced Diffusion Policy's image encoder with a point-cloud encoder, showing that 3D geometric input substantially improves sample efficiency and viewpoint robustness; iDP3~\cite{ze2024idp3} further scaled this approach to humanoid manipulation with egocentric point clouds, but, like DP3, used a standard (non-equivariant) point-cloud backbone. EquivDP3 is, in this lineage, a drop-in replacement of DP3's encoder rather than a change to the diffusion backbone itself, which isolates the effect of equivariance from confounds in the generative model.

\subsection{Equivariant Representations for Manipulation}
Vector Neuron Networks (VNN)~\cite{deng2021vnn} lift scalar neuron activations to $\mathbb{R}^3$-valued ``vector neurons'' and replace standard linear and nonlinear layers with rotation-equivariant counterparts (VN-Linear, VN-LeakyReLU), yielding networks that are equivariant to SO(3) by construction rather than through data augmentation. EquiBot~\cite{zhou2024equibot} combined a VNN point-cloud encoder with a diffusion policy to obtain a controller equivariant to the larger similarity group SIM(3) (rotation, translation, and uniform scale), and demonstrated improved data efficiency and generalization on a wheeled mobile-manipulator platform across several tabletop and mobile manipulation tasks. Equivariant Diffusion Policy~\cite{wang2024equidiff} goes further in a different direction, making the denoising network itself equivariant to planar (SO(2)) rotations so that the action distribution, not only the perception latent, inherits the symmetry. Our work is, to our knowledge, the first to carry this SIM(3)-equivariant encoder into a full humanoid loco-manipulation pipeline, where the action space, embodiment dynamics, and balance constraints are qualitatively different from a wheeled base. We additionally benchmark against two further non-equivariant point-cloud architectures not considered by EquiBot (DGCNN~\cite{wang2019dgcnn}, which builds dynamic local neighborhood graphs, and PCT~\cite{guo2021pct}, a point-cloud transformer with self-attention) alongside PointNet~\cite{qi2017pointnet} and a PointNet variant trained with explicit SO(3) data augmentation, to test whether equivariance provides benefits beyond what either stronger local feature extraction or augmentation already capture.

A broader line of work exploits geometric structure for manipulation without going through a VNN point-cloud encoder. Neural and Equivariant Descriptor Fields~\cite{simeonov2022ndf,ryu2023edf} learn SE(3)-equivariant object representations directly from demonstrations for pick-and-place; EquivAct~\cite{yang2024equivact} extends SIM(3)-equivariant visuomotor policies beyond rigid objects; equivariant transporter networks~\cite{huang2022equivtransporter} bake SO(2)/SE(2) symmetry into pixel-wise pick-place policies for sample-efficient tabletop manipulation, and Fourier Transporter~\cite{huang2024fouriertransporter} extends this to the bi-equivariant (pick and place) symmetry of 3D manipulation; and SE(3)-DiffusionFields~\cite{urain2023se3dif} learns grasp cost functions by diffusion on SE(3) for joint grasp-and-motion optimization. These target fixed-base or mobile tabletop manipulation; none, to our knowledge, has been evaluated on a humanoid whose own base pose is being controlled simultaneously with the manipulation task, which is the setting that motivates the two-stage architecture in Sec.~\ref{sec:method}.

\subsection{Humanoid Whole-Body Control}
A separate line of work addresses how to actually drive a humanoid's many joints once a high-level command is known. OmniH2O~\cite{he2024omnih2o} and HOVER~\cite{he2025hover} learn versatile whole-body controllers via large-scale reinforcement learning that can track diverse high-level commands (joint targets, end-effector poses, velocity commands) for teleoperation and autonomous control. ResMimic~\cite{resmimic2025} uses residual learning to adapt whole-body motion-tracking policies to loco-manipulation. These approaches generally train a single RL whole-body controller end-to-end. We instead build on a stack that keeps the locomotion policy and the manipulation (arm) control as separate low-level modules (a frozen, independently trained RL locomotion policy, HOMIE~\cite{ben2025homie}, driven by simple velocity/angular and torso commands, and a model-based inverse-kinematics solver for the arms, PINK~\cite{caron2024pink}), so that the diffusion planner only needs to reason about task-level intent (where to move the base, where to put the hands) rather than joint-level dynamics. REFINE-DP~\cite{refinedp2026} fine-tunes a diffusion planner with reinforcement learning for humanoid loco-manipulation; our planner is trained by behavior cloning only, and such fine-tuning is complementary to the encoder change we study.

\section{Method}
\label{sec:method}

\subsection{Problem Formulation}
We formulate humanoid loco-manipulation as a partially observed Markov decision process. At each high-level decision step $t$, the observation $\mathcal{O}_t$ consists of an egocentric point cloud $P_t \in \mathbb{R}^{N \times 3}$ ($N{=}1024$ points, obtained by depth unprojection and subsampling) and a proprioceptive vector $p_t \in \mathbb{R}^{28}$ containing the right and left end-effector positions and quaternions, a body/torso end-effector pose, and the robot base position and orientation. The policy outputs a whole-body command chunk $A_t = \{c_t, \dots, c_{t+T_{act}-1}\}$, where each $c_t \in \mathbb{R}^{23}$ packs: left/right hand open/close state (indices 0--1), left wrist position and quaternion (2--8), right wrist position and quaternion (9--15), a planar navigation command $(v_x, v_y, \omega_z)$ (16--18), a base-height command (19), and a torso roll-pitch-yaw command (20--22). 

\subsection{System Overview: A Two-Stage Policy}
\begin{figure*}[t]
\centering
\includegraphics[width=\linewidth]{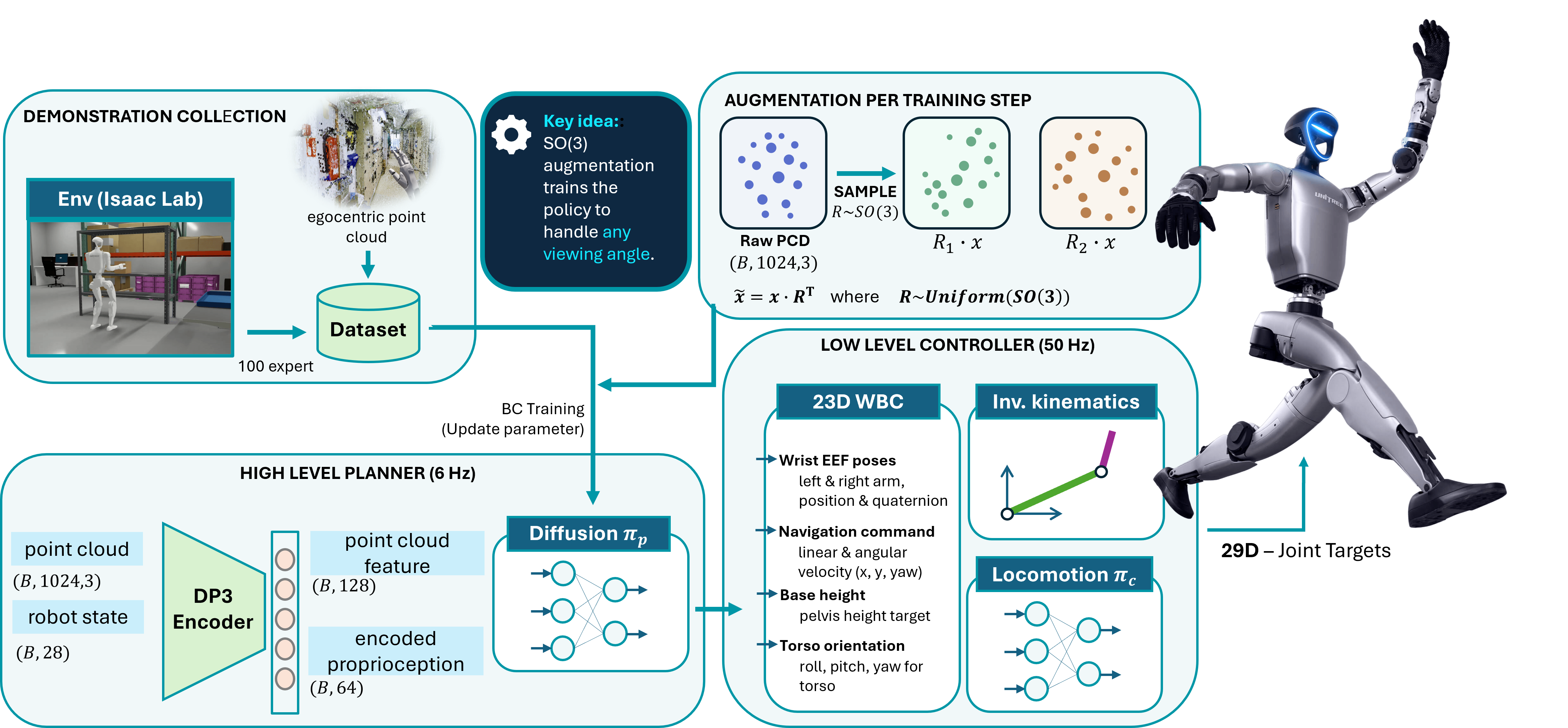}
\caption{EquivDP3 two-stage architecture. A diffusion planner with a SIM(3)-invariant point-cloud encoder runs at $\sim$6\,Hz and emits whole-body command chunks. A low-level controller at 50\,Hz splits each command between an RL-trained locomotion policy (velocity/angular/height/torso $\rightarrow$ leg joints) and an inverse-kinematics module (wrist pose $\rightarrow$ arm joints); both feed a joint-space PD controller.}
\label{fig:system-overview}
\end{figure*}

Fig.~\ref{fig:system-overview} shows the overall architecture. EquivDP3 is split into two stages that run at different control rates, communicating only through a low-dimensional command interface.

\textbf{High-level diffusion planner} ($\sim$6\,Hz). A SIM(3)-invariant point-cloud encoder maps the latest egocentric point cloud to a latent $z_{pc}$; this is concatenated with an embedded proprioception history to form the conditioning vector $o_{cond}$, which is fed to a conditional U-Net diffusion model that generates an 8-step, 23-dimensional whole-body command chunk via DDIM sampling. The planner re-plans every $T_{act}{=}8$ low-level \looseness=-1 steps.

\textbf{Low-level controller} ($\sim$50\,Hz). The 23-dimensional command is split and routed to two functionally separate modules:
\begin{itemize}
\item a \textbf{locomotion policy} (HOMIE~\cite{ben2025homie}), \emph{trained with reinforcement learning} and frozen at deployment, which takes the velocity and angular-velocity command $(v_x, v_y, \omega_z)$ together with the base-height and torso commands as input and outputs leg joint targets that keep the robot balanced while tracking the commanded base motion;
\item an \textbf{inverse-kinematics} module (PINK~\cite{caron2024pink}), which takes the left/right wrist position and quaternion targets and solves for arm joint targets via differential IK.
\end{itemize}
Both joint target streams are tracked by a low-level PD controller at the robot's native control rate. This split is the central architectural property of the stack we build on: \emph{the diffusion planner never directly outputs joint torques or even joint angles for the legs; it only ever issues a velocity/angular-velocity/height/torso command that is interpreted by a pre-trained RL locomotion policy.} This keeps the diffusion planner's output space low-dimensional and largely embodiment-agnostic (a navigation command plus wrist poses), while delegating the hard, dynamics-sensitive problem of bipedal balance to a controller trained for that purpose.

\subsection{SIM(3)-Equivariant Point-Cloud Encoder}
A function $f$ is equivariant to a group $G$ acting on its input and output spaces if $f(g \cdot P) = \rho(g) \cdot f(P)$ for every $g \in G$. We consider $G{=}$SIM(3), whose elements $g{=}(R,t,s)$ act on a point $x \in \mathbb{R}^3$ as $g \cdot x = sRx + t$, with $R \in SO(3)$ a rotation, $t \in \mathbb{R}^3$ a translation, and $s \in \mathbb{R}_{>0}$ a uniform scale. A standard PointNet-style encoder, which applies a shared per-point scalar MLP followed by max-pooling, is invariant to point permutation but has no guarantee of behaving predictably under $g \cdot P$: it must learn this behavior from data, which is the usual argument for why such encoders are data-hungry.

We instead build the encoder out of Vector Neuron layers, which operate on \emph{vector} features $V \in \mathbb{R}^{C\times3}$ (each of the $C$ channels is a 3-vector, not a scalar) so that rotations act naturally by right-multiplication. A VN-Linear layer with weight $W \in \mathbb{R}^{C_{out}\times C_{in}}$ computes
\begin{equation}
\text{VNLinear}(V) = WV, \qquad V \in \mathbb{R}^{C_{in}\times 3},
\label{eq:vnlinear}
\end{equation}
applied identically to each of the 3 spatial columns and with no bias term (a bias would break equivariance of the per-channel direction). Our nonlinearity is a directional \emph{gate}: a learned equivariant direction $q = W_q V$ is computed by mixing channels, and each channel's vector $v$ is scaled by $\alpha$ when its projection onto the corresponding direction is negative:
\begin{equation}
\text{VNGate}(v) =
\begin{cases}
v, & v \cdot q \geq 0 \\
\alpha\, v, & v \cdot q < 0
\end{cases}
\label{eq:vnlrelu}
\end{equation}
with $\alpha{=}0$ in our implementation, i.e.\ a hard gate. This differs from the leaky formulation of Deng et al.~\cite{deng2021vnn}, which subtracts only the component along $\hat q = q/\|q\|$ rather than rescaling the entire vector; both preserve equivariance, since the gating scalar $v \cdot q$ is an inner product of two equivariantly-transformed vectors and is therefore invariant, while $v$ itself transforms correctly. Finally, a VN-StdFeature layer extracts SO(3)-\emph{invariant} scalar features from the equivariant vector features (via inner products with a learned equivariant frame), which is what allows the final latent $z_{pc}$ to be a plain vector consumable by the (non-equivariant) diffusion backbone.

Our encoder stacks four VN-Linear/VN-Gate blocks with hidden widths $[32,64,128,128]$, followed by norm-based max-pooling over the $N$ points (each channel keeps its largest-norm vector), VN-StdFeature, a LayerNorm~\cite{ba2016layernorm} on the resulting invariant descriptor, and a linear layer with a second LayerNorm projecting to the final 128-dimensional latent $z_{pc}$. Algorithm~\ref{alg:encoder} summarizes the forward pass.

Three distinct mechanisms produce the three invariances, and it is worth separating them because they are not equally structural. \emph{Rotation} invariance is the genuine architectural property, guaranteed by the Vector Neuron construction (Eqs.~\ref{eq:vnlinear}--\ref{eq:vnlrelu}) regardless of any normalization choice. \emph{Translation} invariance comes from mean-centering the point cloud before the first layer. \emph{Scale} invariance, however, is not obtained by an explicit normalization step: it emerges from the LayerNorm applied to the invariant descriptor before the output projection, which removes the global multiplicative factor that a uniform rescaling of the input induces. We verified all three empirically (Table~\ref{tab:equivariance}); we note the LayerNorm dependence explicitly because disabling that normalization, which would ordinarily read as a training-stability choice, silently removes scale invariance while leaving rotation and translation intact.

\begin{algorithm}[t]
\caption{SIM(3)-Equivariant Point-Cloud Encoding}
\label{alg:encoder}
\begin{algorithmic}[1]
\REQUIRE point cloud $P \in \mathbb{R}^{N\times 3}$, $N{=}1024$
\STATE $\bar P \leftarrow P - \text{mean}(P)$ \hfill {\small // translation-invariant centering}
\STATE $F_0 \leftarrow \bar P$ \hfill {\small // $N$ vector-features, each in $\mathbb{R}^{1\times 3}$}
\FOR{$\ell = 1$ to $4$}
\STATE $F_\ell \leftarrow \text{VNGate}(\text{VNLinear}_\ell(F_{\ell-1}))$
\ENDFOR
\STATE \hfill {\small // hidden dims: 32, 64, 128, 128}
\STATE $F_{\text{glob}} \leftarrow \text{MaxPool}_N(F_4)$ \hfill {\small // per-channel argmax by norm}
\STATE $F_{\text{inv}} \leftarrow \text{VNStdFeature}(F_{\text{glob}})$ \hfill {\small // invariant scalar features}
\STATE $z_{pc} \leftarrow \text{LN}(\text{Linear}(\text{LN}(F_{\text{inv}}))) \in \mathbb{R}^{128}$ \hfill {\small // LN: LayerNorm}
\ENSURE $z_{pc}$ \hfill {\small // rotation-, translation-, and scale-invariant latent}
\end{algorithmic}
\end{algorithm}

\begin{table}[t]
\centering
\caption{Empirical invariance check. Relative error $\|f(g\!\cdot\!P) - f(P)\|/\|f(P)\|$ over random transforms; $10^{-7}$ is float32 round-off. The final row shows that disabling the encoder's LayerNorms removes scale invariance only.}
\label{tab:equivariance}
\begin{threeparttable}
\begin{tabular}{lccc}
\toprule
\textbf{Encoder} & \textbf{Rotation} & \textbf{Translation} & \textbf{Scale} \\
\midrule
EquivDP3 (ours)      & $3.9\!\times\!10^{-7}$ & $2.7\!\times\!10^{-7}$ & $3.8\!\times\!10^{-7}$ \\
PointNet             & $2.3\!\times\!10^{-1}$ & $5.3\!\times\!10^{-1}$ & $1.5\!\times\!10^{-1}$ \\
\midrule
Ours, no LayerNorm   & $4.1\!\times\!10^{-7}$ & $6.0\!\times\!10^{-7}$ & $5.2$ \\
\bottomrule
\end{tabular}
\end{threeparttable}
\end{table}

\subsection{High-Level Diffusion Planner}
Given $z_{pc}$ and a 2-step proprioception history, the conditioning vector is formed by concatenating the point-cloud latent with a learned proprioception embedding at each of the $T_{obs}{=}2$ observation steps and flattening,
\begin{equation}
o_{cond} = \big[\,[z_{pc}^{(\tau)};\,\text{MLP}_{s}(p_\tau)]\,\big]_{\tau = t-1}^{t} \in \mathbb{R}^{384},
\label{eq:ocond}
\end{equation}
where $\text{MLP}_{s}: \mathbb{R}^{28} \!\to\! \mathbb{R}^{64}$ is a two-layer state encoder, giving $(128{+}64)\times 2 = 384$ dimensions. This is fed to a conditional U-Net (down-sampling channel widths $[512,1024,2048]$, kernel size 5, 8 GroupNorm groups) via Feature-wise Linear Modulation (FiLM)~\cite{perez2018film}:
\begin{equation}
\begin{gathered}
\text{FiLM}(h;\gamma,\beta) = \gamma \odot h + \beta,\\
(\gamma,\beta) = \text{MLP}_{film}(o_{cond}, \phi(k)),
\end{gathered}
\label{eq:film}
\end{equation}
where $\phi(k)$ is a sinusoidal embedding (dimension 256) of the diffusion timestep $k$. Training follows DDPM~\cite{ho2020ddpm}: the forward (noising) process is
\begin{equation}
q(A_k \mid A_0) = \mathcal{N}\!\left(A_k;\ \sqrt{\bar\alpha_k}\,A_0,\ (1-\bar\alpha_k)\,I\right),
\label{eq:forward}
\end{equation}
with a squared-cosine (\texttt{squaredcos\_cap\_v2}) noise schedule and $K_{train}{=}100$ training steps, and the network $\hat A_\theta$ is trained to directly predict the clean action chunk,
\begin{equation}
\mathcal{L}_{BC}(\theta) = \mathbb{E}_{t,k,\epsilon}\Big[\big\|A_0 - \hat A_\theta(A_k, k, o_{cond})\big\|_2^2\Big].
\label{eq:bc-loss}
\end{equation}
At inference time we use deterministic DDIM sampling~\cite{song2021ddim} with $K_{infer}{=}10$ steps (the $\sigma_k{=}0$ special case of the general DDIM update),
\begin{equation}
\begin{split}
A_{k-1} = {}&\sqrt{\bar\alpha_{k-1}}\,\hat A_\theta(A_k,k,o_{cond})\\
&+ \sqrt{1-\bar\alpha_{k-1}}\cdot \frac{A_k - \sqrt{\bar\alpha_k}\,\hat A_\theta}{\sqrt{1-\bar\alpha_k}},
\end{split}
\label{eq:ddim}
\end{equation}
which reduces sampling from 100 to 10 network evaluations per replanning step, keeping the high-level loop within its $\sim$6\,Hz budget.

\subsection{Low-Level Controller: RL Locomotion Policy and Inverse Kinematics}
\textbf{Locomotion policy (reinforcement-learned).} The leg joints are never commanded directly by the diffusion planner. Instead, a separate locomotion policy $\pi_{loco}$, pre-trained with reinforcement learning in massively parallel simulation by the HOMIE authors~\cite{ben2025homie} (we use the \texttt{homie\_v2} checkpoint shipped with IsaacLab Arena~\cite{isaaclabarena2025}, unmodified) and kept frozen during both imitation-learning training and deployment of the high-level planner, is queried at 50\,Hz with the robot's proprioceptive state $s_t$, the navigation command $(v_x,v_y,\omega_z)$ and the base-height/torso command extracted from $c_t$, and outputs target leg joint angles:
\begin{equation}
q^{\text{leg}}_{tgt} = \pi_{loco}(s_t;\, v_x, v_y, \omega_z, h_{base}, \theta_{torso}).
\label{eq:loco}
\end{equation}
Because $\pi_{loco}$ was trained with RL against the full nonlinear contact dynamics of bipedal walking, it can absorb disturbances and maintain balance in a way that a diffusion model trained purely on a few hundred demonstration trajectories could not be expected to. This division of labor (\emph{diffusion planner decides where to go and what to grasp; an RL locomotion policy decides how to keep the robot upright while getting there}) is, in our view, what makes a hierarchical design preferable to a single end-to-end policy for a humanoid \looseness=-1 embodiment.

\textbf{Inverse kinematics (model-based).} The arm joints are driven by a differential inverse-kinematics solver (PINK~\cite{caron2024pink}) that takes the commanded left/right wrist positions and quaternions and resolves them into arm joint targets $q^{\text{arm}}_{tgt}$ consistent with the G1's kinematic chain, subject to joint limits. Both joint streams are tracked with a standard PD law, $\tau_t = K_p(q_{tgt}-q_t) + K_d(\dot q_{tgt} - \dot q_t)$. Algorithm~\ref{alg:deploy} summarizes the full dual-rate deployment loop, with the RL locomotion policy call explicitly \looseness=-1 marked.

\begin{algorithm}[t]
\caption{Dual-Rate Deployment Loop}
\label{alg:deploy}
\begin{algorithmic}[1]
\STATE \textbf{High-level thread ($\sim$6\,Hz):}
\WHILE{task not done}
\STATE capture depth frame; unproject + subsample $\rightarrow P_t$
\STATE read proprioception $\rightarrow p_t$
\STATE $z_{pc} \leftarrow$ EquivEncoder($P_t$) \hfill {\small // Alg.~\ref{alg:encoder}}
\STATE $o_{cond} \leftarrow [z_{pc};\,\text{MLP}_{s}(p_{t-1:t})]$ \hfill {\small // Eq.~\ref{eq:ocond}}
\STATE $A_t \leftarrow \text{DDIM}_{10}(o_{cond})$ \hfill {\small // 8$\times$23 action chunk}
\STATE write $A_t$ to shared ring buffer
\ENDWHILE
\STATE
\STATE \textbf{Low-level thread (50\,Hz):}
\WHILE{task not done}
\STATE $c_\tau \leftarrow$ next buffered command (hold last if empty)
\STATE $c_\tau[0], c_\tau[1] \leftarrow 0$ \hfill {\small // hands held open (Task 1 only)}
\STATE $q^{\text{arm}}_{tgt} \leftarrow \text{PINK-IK}(c_\tau[2{:}16])$
\STATE $q^{\text{leg}}_{tgt} \leftarrow \pi_{loco}^{\text{RL}}(c_\tau[16{:}23])$ \hfill {\small // RL-trained; $v_x,v_y,\omega_z$ + height/torso}
\STATE $\tau_t \leftarrow K_p(q_{tgt}-q_t) + K_d(\dot q_{tgt}-\dot q_t)$
\STATE apply $\tau_t$ to robot
\ENDWHILE
\end{algorithmic}
\end{algorithm}

Algorithm~\ref{alg:deploy} is written as the loop would run on hardware; in simulation the two rates are interleaved synchronously (one planner call per eight control steps). On hardware, holding the last command when a cycle is missed (line 13) would isolate balance and gait from a stall in the slower, vision-dependent planner, another consequence of delegating balance to the always-on RL locomotion policy.

\subsection{Training: Behavior Cloning}
EquivDP3's high-level planner (invariant encoder + diffusion U-Net) is trained end-to-end using the loss in Eq.~\eqref{eq:bc-loss}, optimized with AdamW~\cite{loshchilov2019adamw} ($\alpha{=}10^{-4}$, $\beta_1{=}0.95$, $\beta_2{=}0.999$, weight decay $10^{-6}$), with an exponential moving average (EMA) of the weights, $\theta_{EMA} \leftarrow \beta_{EMA}\,\theta_{EMA} + (1-\beta_{EMA})\,\theta$, $\beta_{EMA}\rightarrow0.9999$ after warmup, used for evaluation. The locomotion policy $\pi_{loco}$ is \emph{not} part of this stage; it is a separately pre-trained, frozen RL component supplied by the low-level controller, consistent with Section~\ref{sec:method}-E. All results in this paper use behavior cloning only; online RL fine-tuning of the planner (e.g.\ via DPPO~\cite{ren2024dppo}) is compatible with this architecture but is left to future work, since it is orthogonal to the encoder comparison that is our subject.

\textbf{Demonstrations.} Both datasets are generated with IsaacLab Mimic, an implementation of MimicGen~\cite{mandlekar2023mimicgen}, from a small number of annotated human source demonstrations: 200 episodes from a single source demonstration for Task 1 and 360 episodes from five for Task 2, with 10\% held out for validation. A budget of $n_{demo}$ demonstrations is a random subset of the training split (seed 42), drawn separately for each budget, so smaller budgets are not nested in larger ones; for a given task and budget, every encoder trains on the identical subset. All encoders are trained for 100 epochs with seed 42.

\section{Experiments and Results}
\label{sec:experiments}

\subsection{Simulation Platform and Tasks}
\begin{figure*}[t]
\centering
\includegraphics[width=\linewidth]{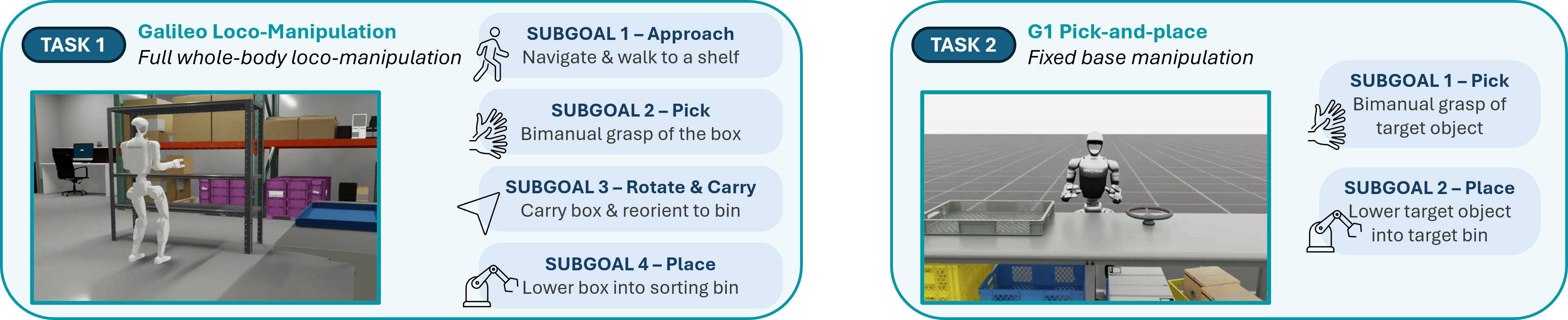}
\caption{Overview of the two evaluation tasks. Task 1 (Galileo Loco-Manipulation, left) requires full whole-body coordination across four subgoals: approach/walk to the shelf, bimanual pick, rotate \& carry, and place into the sorting bin. Task 2 (G1 Pick-and-Place, right) fixes the base and requires only two subgoals: bimanual pick and place into the target bin.}
\label{fig:task-overview}
\end{figure*}

All experiments are run in IsaacLab~\cite{mittal2023orbit,mittal2025isaaclab} with a 43-joint Unitree G1 ``Galileo'' variant (14 finger joints locked for our tasks, leaving 29 actuated body joints). The environment, the frozen locomotion policy and the PINK IK solver are existing IsaacLab Arena~\cite{isaaclabarena2025} infrastructure; our contribution is the encoder and its integration as the planner's perception branch, not the low-level stack. We evaluate on two tasks of differing complexity, illustrated in Fig.~\ref{fig:task-overview} and summarized in Table~\ref{tab:tasks}.

\begin{table}[t]
\centering
\caption{Evaluation tasks.}
\label{tab:tasks}
\begin{threeparttable}
\begin{tabular}{lp{0.4\linewidth}c}
\toprule
\textbf{Task} & \textbf{Subgoals} & \textbf{\# subgoals} \\
\midrule
Galileo Loco-Manip. & approach/walk to shelf $\rightarrow$ bimanual pick $\rightarrow$ rotate \& carry $\rightarrow$ place into bin & 4 \\
G1 Pick-and-Place & pick $\rightarrow$ place (fixed base) & 2 \\
\bottomrule
\end{tabular}
\end{threeparttable}
\end{table}

Task 1 exercises the locomotion policy and the manipulation pipeline together; Task 2 fixes the base, isolating manipulation from locomotion and serving as a simpler control condition.

We evaluate two distribution conditions: \textbf{in-distribution (ID)}, matching the training object pose and lighting distribution, and \textbf{out-of-distribution (OOD)}, where object yaw is randomized beyond the training range and scene lighting (dome light intensity and RGB tint) is randomized. We train and evaluate at four demonstration budgets, $n_{demo} \in \{5,10,50,100\}$, to characterize data efficiency.

\subsection{Baselines}
We compare our SIM(3)-invariant VNN encoder (EquivDP3) against four point-cloud encoder baselines, all otherwise sharing the same DP3 diffusion backbone, training procedure, and hyperparameters: \textbf{PointNet}~\cite{qi2017pointnet} (the original DP3 encoder, no equivariance); \textbf{PointNet+Aug}, the same architecture trained with explicit SO(3) rotation data augmentation, to test whether augmentation alone closes the gap to a built-in equivariance guarantee; \textbf{DGCNN}~\cite{wang2019dgcnn}, which builds dynamic local neighborhood graphs and is a stronger local-feature baseline; and \textbf{PCT}~\cite{guo2021pct}, a point-cloud transformer using self-attention over points.

\subsection{Data Efficiency}
\begin{figure*}[t]
\centering
\includegraphics[width=\linewidth]{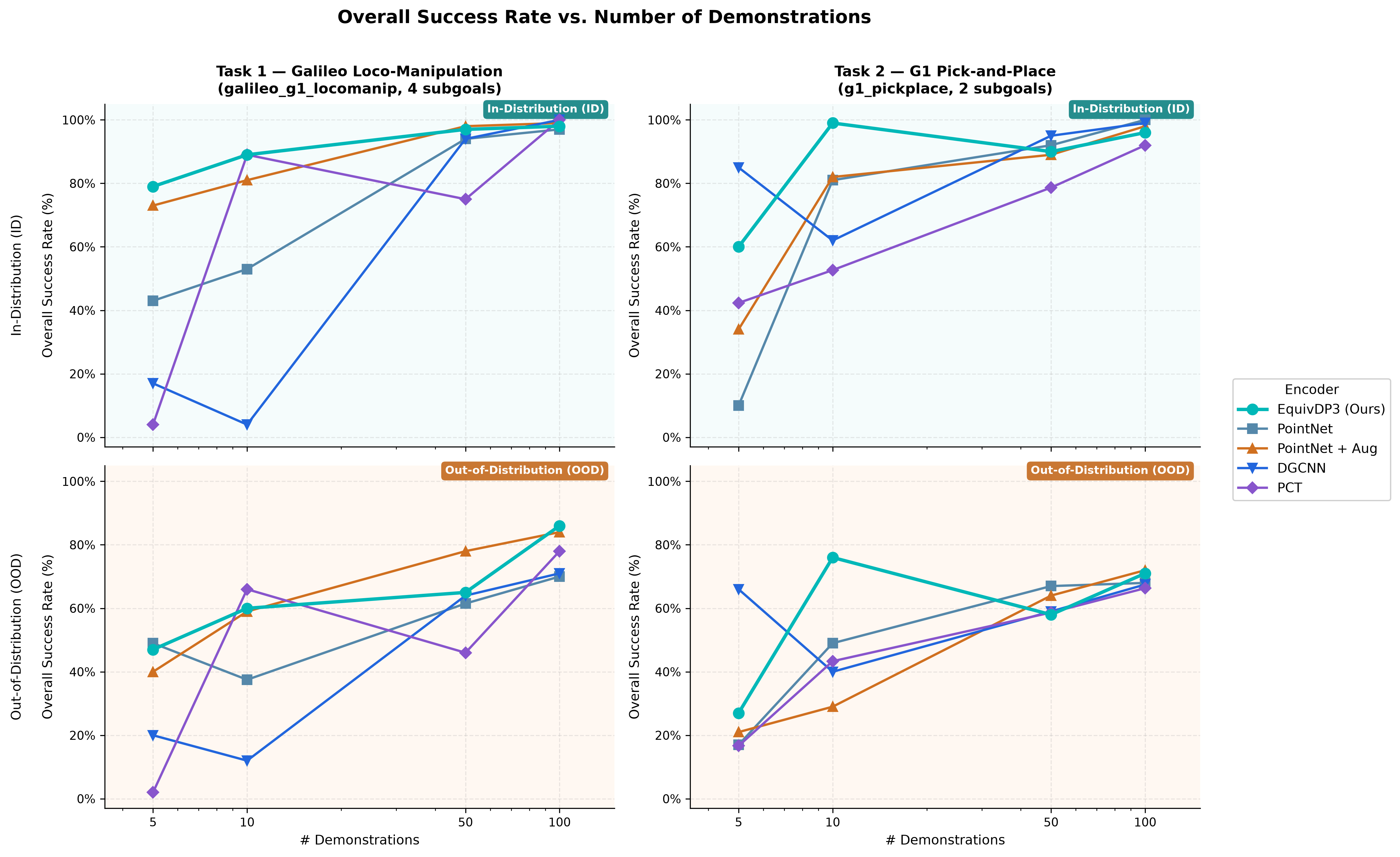}
\caption{Overall success rate vs. number of demonstrations, for Task 1 (Galileo Loco-Manipulation, left column) and Task 2 (G1 Pick-and-Place, right column), under in-distribution (top row) and out-of-distribution (bottom row) conditions. EquivDP3 (teal) leads in aggregate at 5--10 demonstrations; by 50--100 all encoders converge, and it is not the top performer in every individual setting.}
\label{fig:main-results}
\end{figure*}

Fig.~\ref{fig:main-results} reports overall task success rate against the number of demonstrations for both tasks and both distribution conditions; Table~\ref{tab:main-results} gives the corresponding numeric values from the raw logged evaluation results.

\begin{table*}[t]
\centering
\caption{Success rate (\%) by encoder, task, and demonstration count, with in-distribution (ID) and out-of-distribution (OOD) as paired sub-columns under each encoder. Each cell reports the point estimate and the half-width of its 95\% Wilson score interval~\cite{wilson1927} ($p\pm h$), computed directly from the logged per-episode outcomes. Episodes per cell: 63 cells use 100, 6 use 150, 9 use 200, 2 use 300. Cells evaluated more than once on the identical checkpoint are pooled, with $n$ the pooled count; which cells were repeated reflects development history, not selection on outcome.}
\label{tab:main-results}
\small
\setlength{\tabcolsep}{1.5pt}
\begin{tabular}{ll cc cc cc cc cc}
\toprule
& & \multicolumn{2}{c}{\textbf{EquivDP3}} & \multicolumn{2}{c}{\textbf{PointNet}} & \multicolumn{2}{c}{\textbf{PointNet+Aug}} & \multicolumn{2}{c}{\textbf{DGCNN}} & \multicolumn{2}{c}{\textbf{PCT}} \\
\cmidrule(lr){3-4}\cmidrule(lr){5-6}\cmidrule(lr){7-8}\cmidrule(lr){9-10}\cmidrule(lr){11-12}
\textbf{Task} & \textbf{$n_{demo}$} & ID & OOD & ID & OOD & ID & OOD & ID & OOD & ID & OOD \\
\midrule
\multirow{4}{*}{Task 1}
& 5   & \cellpmb{79.00}{7.91} & \cellpm{47.00}{9.60} & \cellpm{43.00}{9.53} & \cellpmb{49.00}{9.61} & \cellpm{73.00}{8.58} & \cellpm{40.00}{9.43} & \cellpm{17.00}{7.33} & \cellpm{20.00}{7.77} & \cellpm{4.00}{4.14} & \cellpm{2.00}{3.23} \\
& 10  & \cellpmb{89.00}{6.19} & \cellpm{60.00}{9.43} & \cellpm{53.00}{9.60} & \cellpm{37.50}{6.65} & \cellpm{81.00}{7.63} & \cellpm{59.00}{9.47} & \cellpm{4.00}{4.14} & \cellpm{12.00}{6.41} & \cellpmb{89.00}{6.19} & \cellpmb{66.00}{9.13} \\
& 50  & \cellpm{97.00}{3.71} & \cellpm{65.00}{9.19} & \cellpm{94.00}{4.85} & \cellpm{61.50}{6.68} & \cellpmb{98.00}{3.23} & \cellpmb{78.00}{8.03} & \cellpm{94.00}{4.85} & \cellpm{64.00}{9.25} & \cellpm{75.00}{5.96} & \cellpm{46.00}{9.59} \\
& 100 & \cellpm{98.00}{3.23} & \cellpmb{86.00}{6.81} & \cellpm{97.00}{3.71} & \cellpm{70.00}{6.30} & \cellpm{99.00}{2.64} & \cellpm{84.00}{7.16} & \cellpmb{100.00}{1.85} & \cellpm{71.00}{8.76} & \cellpmb{100.00}{1.85} & \cellpm{78.00}{8.03} \\
\midrule
\multirow{4}{*}{Task 2}
& 5   & \cellpm{60.00}{9.43} & \cellpm{27.00}{8.58} & \cellpm{10.00}{4.19} & \cellpm{17.00}{7.33} & \cellpm{34.00}{6.51} & \cellpm{21.00}{7.91} & \cellpmb{85.00}{6.99} & \cellpmb{66.00}{9.13} & \cellpm{42.33}{5.56} & \cellpm{16.67}{5.95} \\
& 10  & \cellpmb{99.00}{1.65} & \cellpmb{76.00}{8.27} & \cellpm{81.00}{7.63} & \cellpm{49.00}{9.61} & \cellpm{82.00}{7.48} & \cellpm{29.00}{8.76} & \cellpm{62.00}{9.35} & \cellpm{40.00}{9.43} & \cellpm{52.67}{7.89} & \cellpm{43.33}{7.83} \\
& 50  & \cellpm{90.00}{5.96} & \cellpm{58.00}{9.50} & \cellpm{92.00}{5.44} & \cellpmb{67.00}{9.07} & \cellpm{89.00}{6.19} & \cellpm{64.00}{9.25} & \cellpmb{95.00}{4.51} & \cellpm{59.00}{9.47} & \cellpm{78.67}{6.51} & \cellpm{58.67}{7.78} \\
& 100 & \cellpm{96.00}{4.14} & \cellpm{71.00}{8.76} & \cellpmb{100.00}{1.85} & \cellpm{68.00}{9.00} & \cellpm{98.00}{3.23} & \cellpmb{72.00}{8.67} & \cellpm{99.00}{1.65} & \cellpm{67.50}{6.44} & \cellpm{92.00}{4.41} & \cellpm{66.33}{5.32} \\
\bottomrule
\end{tabular}
\end{table*}

The headline pattern is a regime split. Aggregated over both tasks and distribution conditions, EquivDP3 leads when demonstrations are scarce, not when they are plentiful:

\begin{center}
\small
\begin{tabular}{lcc}
\toprule
\textbf{Encoder} & \textbf{$n_{demo}\!\in\!\{5,10\}$} & \textbf{$n_{demo}\!\in\!\{50,100\}$} \\
\midrule
EquivDP3 (ours) & \textbf{67.1} & 82.6 \\
PointNet+Aug    & 52.4 & \textbf{85.2} \\
PointNet        & 42.4 & 81.2 \\
PCT             & 39.5 & 74.3 \\
DGCNN           & 38.2 & 81.2 \\
\bottomrule
\end{tabular}
\end{center}

At 5--10 demonstrations EquivDP3 leads the best baseline by 14.7 points; at 50--100 it is second, 2.6 points behind PointNet+Aug. We report both halves because the second is as informative as the first: whatever the equivariant encoder provides, it is not a uniform improvement, and a grand mean over all demonstration budgets would obscure exactly that.

Individual cells in the low-data regime are volatile, and we do not lean on any single one. On Task 1 ID, EquivDP3 reaches 79\% and 89\% at $n{=}5$ and $n{=}10$ against 43--73\% and 53--81\% for the PointNet variants, but PCT swings from 4\% at $n{=}5$ to 89\% at $n{=}10$, and DGCNN collapses to 4\% at $n{=}10$ having scored 17\% at $n{=}5$. On Task 2 at $n{=}5$, EquivDP3 trails DGCNN outright (60\% vs.\ 85\% ID), a lead DGCNN then loses at $n{=}10$ (62\%) before regaining at $n{=}50$. These are not stable rankings; the aggregate above is the level at which we consider the comparison meaningful, and Sec.~\ref{sec:experiments}-F quantifies why.

\textbf{Where the advantage comes from.} To locate the low-data advantage within the task, we decompose Task 1 into its four subgoal events (Table~\ref{tab:subgoals}). SG1 fires when any wrist link comes within 0.45\,m of the box; SG2 when the box rises more than 0.05\,m above its start height; SG3 when the box comes within 0.60\,m of the bin, and only after SG2; SG4 is task success as reported by the environment's termination criterion, and is \emph{not} conditioned on SG2. SG2 is therefore a lift detector rather than a grasp detector, and rates need not decrease left to right: an episode that slides the box into the bin without clearing the lift threshold registers SG4 but not \looseness=-1 SG2.

\begin{table}[t]
\centering
\caption{Task 1 subgoal event rates (\%), ID, at the two low-data budgets. SG1 approach, SG2 lift, SG3 carry to bin (gated on SG2), SG4 task success (not gated on SG2).}
\label{tab:subgoals}
\small
\setlength{\tabcolsep}{3.5pt}
\begin{tabular}{lcccc cccc}
\toprule
& \multicolumn{4}{c}{$n_{demo}{=}5$} & \multicolumn{4}{c}{$n_{demo}{=}10$} \\
\cmidrule(lr){2-5}\cmidrule(lr){6-9}
\textbf{Encoder} & SG1 & SG2 & SG3 & SG4 & SG1 & SG2 & SG3 & SG4 \\
\midrule
EquivDP3     & 100 & \textbf{90} & \textbf{87} & \textbf{79} & 100 & 98 & 97 & \textbf{89} \\
PointNet     & 100 & 32 & 27 & 43 & 100 & 79 & 53 & 53 \\
PointNet+Aug & 100 & 77 & 76 & 73 & 100 & 99 & 98 & 81 \\
DGCNN        & 100 & 42 & 29 & 17 & 100 & \textbf{100} & \textbf{100} & 4 \\
PCT          & 100 & 18 & 17 & 4 & 100 & 99 & 96 & \textbf{89} \\
\bottomrule
\end{tabular}
\end{table}

At $n_{demo}{=}5$ every encoder completes the approach subgoal on every episode: walking to the shelf is executed by the frozen locomotion policy, which never consumes $z_{pc}$, so no encoder can distinguish itself there. The five-way spread opens at the lift, the first step that requires the box to be acquired and raised (90\% for EquivDP3 against 77, 42, 32 and 18\% for the baselines), and the ordering set there largely persists to the end of the episode. The low-data advantage is therefore localized to the phase that requires inferring object geometry from the point cloud, and is absent from the phase the two-stage architecture delegates to a separately trained controller, which is direct evidence for the separation described in Sec.~\ref{sec:method}. Because SG2 is a lower bound on acquisition (PointNet at $n_{demo}{=}5$ succeeds on 43\% of episodes while registering a lift on 32\%), the localization should be read at the granularity of the manipulation phase rather than of a verified grasp. By $n_{demo}{=}10$ the lift gap has closed for every encoder but PointNet, matching the regime split above; DGCNN then reaches the bin on every episode yet completes the task on 4\%. The same holds on Task 2: at $n_{demo}{=}5$ EquivDP3 has the highest lift rate of any encoder (79\% against 43--68\%).

\subsection{Out-of-Distribution Robustness}
\begin{figure*}[t]
\centering
\begin{subfigure}[b]{0.49\linewidth}
\includegraphics[width=\linewidth]{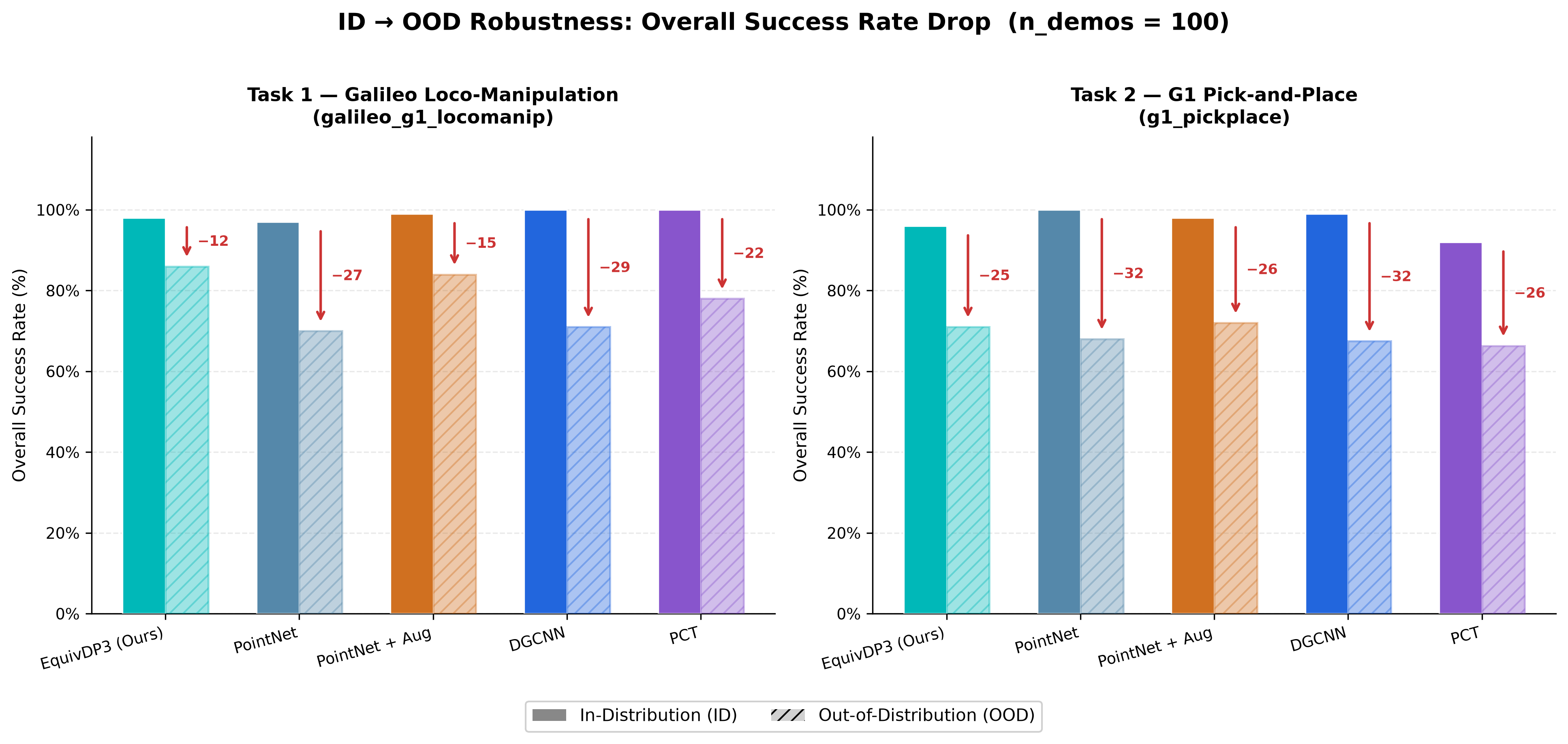}
\caption{$n_{demo}=100$}
\label{fig:ood-drop-100}
\end{subfigure}
\hfill
\begin{subfigure}[b]{0.49\linewidth}
\includegraphics[width=\linewidth]{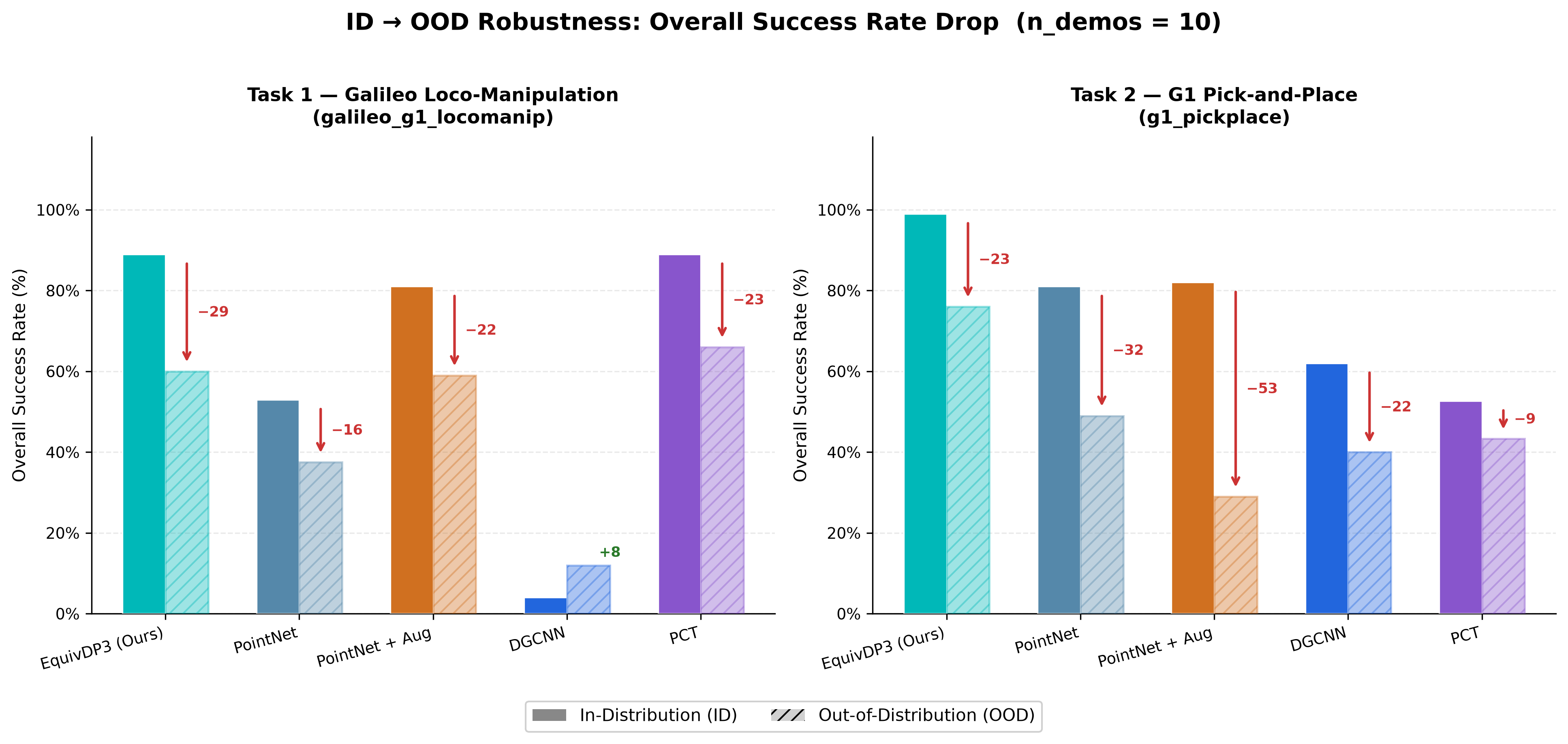}
\caption{$n_{demo}=10$}
\label{fig:ood-drop-10}
\end{subfigure}
\caption{ID$\rightarrow$OOD success-rate drop (percentage points) for both tasks, at (a) $n_{demo}{=}100$ and (b) $n_{demo}{=}10$. EquivDP3 shows the smallest drop among all encoders at $n{=}100$ on Task 1, and DGCNN nearly collapses on Task 1 at $n{=}10$ (both ID and OOD performance near floor).}
\label{fig:ood-drop}
\end{figure*}

Under OOD shift the picture is less clear. Fig.~\ref{fig:ood-drop} shows the ID-to-OOD success-rate drop at $n_{demo}{=}100$ (panel a, the well-resourced regime) and at $n_{demo}{=}10$ (panel b, the low-data regime), for both tasks.

At $n_{demo}{=}100$ on Task 1, EquivDP3 shows the smallest ID-to-OOD drop of any encoder: 12 percentage points (98\%$\rightarrow$86\%), against 15 (PointNet+Aug), 22 (PCT), 27 (PointNet), and 29 (DGCNN). We avoid a stronger summary than that. Against the \emph{best-performing} baseline on this metric the margin is 3 points, which is smaller than the run-to-run spread we measure elsewhere (Sec.~\ref{sec:experiments}-F), and the pattern does not hold uniformly: across the eight (task, $n_{demo}$) settings we evaluate, EquivDP3 has the smallest drop in only two. At $n_{demo}{=}50$ on Task 1 it is fourth of five. We therefore report the $n_{demo}{=}100$ Task 1 result as a single favorable observation rather than as a general robustness claim; the data-efficiency result of Sec.~\ref{sec:experiments}-C, which is both larger in magnitude and consistent across tasks, is the finding we stand \looseness=-1 behind.

At $n_{demo}{=}10$ the picture is noisier still. DGCNN's near-collapse on Task 1 (4\% ID, 12\% OOD) is a different failure mode from a poor ID-to-OOD transfer, and the subgoal decomposition locates it: in distribution DGCNN lifts the box and carries it to the bin on every episode, then fails the final placement (Table~\ref{tab:subgoals}). Note also that several $n_{demo}{=}5$ cells show \emph{negative} drops (OOD scoring above ID), a direct signature of noise dominating at that sample size, and a further reason to treat individual OOD cells cautiously. The proprioception-only control of Sec.~\ref{sec:experiments}-E sharpens this caution considerably. Run under the same OOD condition on Task 2, a policy with no visual input is statistically indistinguishable from every encoder at $n_{demo}{\geq}50$, and significantly \emph{outperforms} three of the five at $n_{demo}{=}10$. Much of the Task 2 OOD column therefore measures something other than perception, and ID-to-OOD drops computed from it should not be read as a ranking of encoder robustness. We have no blind control on Task 1, where the drop discussed above was measured, so we cannot say whether the same holds there.

\subsection{Is Vision Necessary? A Proprioception-Only Control}
\label{sec:blind}

A success-rate comparison between point-cloud encoders is only meaningful if the benchmark requires the point cloud. To test this directly we train and evaluate a \textbf{blind} control: the identical policy, architecture, training schedule, and data, with the point-cloud observation zeroed inside the encoder, so that no scene information reaches the planner and it must act from proprioception alone. Everything else is unchanged: the encoder still runs, it simply receives a constant input. Under ID the object pose matches training, so a policy that has effectively memorized one trajectory can still succeed; that is the ceiling this control is meant to expose. We run it under both distribution conditions.

\begin{table}[t]
\centering
\caption{Proprioception-only control on Task 2 across the full demonstration-budget grid of Table~\ref{tab:main-results}, under both distribution conditions. Bold gaps are significant by Fisher's exact test ($p{<}0.05$).}
\label{tab:blind}
\begin{threeparttable}
\begin{tabular}{lcccc}
\toprule
& \textbf{$n{=}5$} & \textbf{$n{=}10$} & \textbf{$n{=}50$} & \textbf{$n{=}100$} \\
\midrule
\multicolumn{5}{l}{\emph{In-distribution}} \\
Blind (no point cloud)\tnote{a} & 31 & 78 & 92 & 94 \\
EquivDP3 (ours)                 & 60 & 99 & 90 & 96 \\
Gap (EquivDP3 $-$ Blind)        & \textbf{+29} & \textbf{+21} & $-2$ & $+2$ \\
\midrule
\multicolumn{5}{l}{\emph{Out-of-distribution}} \\
Blind (no point cloud)\tnote{a} & 21 & 62 & 65 & 64 \\
EquivDP3 (ours)                 & 27 & 76 & 58 & 71 \\
Gap (EquivDP3 $-$ Blind)        & $+6$ & \textbf{+14} & $-7$ & $+7$ \\
\bottomrule
\end{tabular}
\begin{tablenotes}
\small
\item[a] 100 evaluation episodes, matching EquivDP3; single seed, single training run per $n_{demo}$.
\end{tablenotes}
\end{threeparttable}
\end{table}

The result (Table~\ref{tab:blind}) is a gradient rather than a step function. The perceptual gap is largest at $n_{demo}{=}5$ (29 points, blind 31\% vs.\ EquivDP3 60\%) and somewhat smaller at $n_{demo}{=}10$ (21 points, 78\% vs.\ 99\%); both are significant by Fisher's exact test ($p{<}10^{-4}$ each), so vision remains necessary at both budgets that make up our low-data claim. By $n_{demo}{=}50$ the gap has closed and nominally inverts (92\% vs.\ 90\%, $p{=}0.81$), and at $n_{demo}{=}100$ it remains within noise (94\% vs.\ 96\%, $p{=}0.75$); neither difference is separable from sampling variation. The benchmark's perceptual content is therefore concentrated below $n_{demo}{=}50$, with the transition occurring between 10 and 50 demonstrations rather than at a single \looseness=-1 threshold.

Under distribution shift the blind policy loses ground, as it should: randomized yaw breaks a memorized trajectory, and blind success falls by 10 to 30 points relative to ID. It does not, however, collapse. From $n_{demo}{=}10$ upward it succeeds on 62--65\% of episodes, and against it the five encoders fare as follows (Fisher's exact test, $p{<}0.05$). At $n_{demo}{=}5$ only DGCNN significantly outperforms the blind policy (66\% vs.\ 21\%). At $n_{demo}{=}10$ only EquivDP3 does (76\% vs.\ 62\%, $p{=}0.046$), while PointNet+Aug (29\%), DGCNN (40\%) and PCT (43\%) are significantly \emph{worse} than a policy that cannot see, which suggests that under distribution shift their visual features mislead the planner more than they inform it. At $n_{demo}{=}50$ and $100$ no encoder is distinguishable from the blind policy. The ceiling is therefore not specific to in-distribution evaluation: under the OOD condition we use, the Task 2 comparison is informative about perception only at the smallest budgets.

We draw three conclusions. First, the data-efficiency result of Sec.~\ref{sec:experiments}-C is a genuine perceptual effect at both $n_{demo}{=}5$ and $n_{demo}{=}10$ in distribution: removing vision costs EquivDP3 29 and 21 points in the regime where we claim an advantage. Second, cells at $n_{demo}{\geq}50$ should not be read as evidence about encoders under \emph{either} distribution condition. Third, widening the placement distribution, as our OOD condition does for object yaw, is not by itself enough to make a benchmark perceptually demanding: a blind policy still succeeds on roughly two episodes in three. We report this control because it is, to our knowledge, the most direct available check on whether an encoder comparison is measuring what it claims to measure, and it is inexpensive to run.

\subsection{Reliability of Individual Cells}
\label{sec:reliability}

Each cell in Table~\ref{tab:main-results} is a single training run, evaluated on 100 episodes except where repeated evaluations of the identical checkpoint were pooled (63 cells use 100 episodes, 6 use 150, 9 use 200 and 2 use 300). Which cells were repeated reflects development history, not selection on outcome, and every result uses the final training checkpoint: no checkpoint is selected on evaluation success. Two sources of variance are therefore unmeasured in the headline table, and we quantify them here rather than leave them implicit.

\textbf{Evaluation variance.} Several cells were evaluated more than once from the identical checkpoint during the course of this work. The resulting spreads are substantial: 9 points (PointNet, Task 1 OOD, $n{=}10$: 33\% vs.\ 42\%), 8 points (PointNet, Task 2 ID, $n{=}5$: 6\% vs.\ 14\%), 7 points (PointNet, Task 1 OOD, $n{=}50$: 58\% vs.\ 65\%), and 10 points even at three evaluation seeds and 150 episodes (PCT, Task 2 ID, $n{=}5$: 37.3\% vs.\ 47.3\%). Differences below roughly 10 points between individual cells should therefore not be interpreted.

\textbf{Training variance.} No (task, budget) configuration is trained more than once, so the variance induced by re-training is not measured cell by cell, and it is additional to the above. The low-data aggregate is nonetheless not a single run: ID and OOD share a checkpoint, so it averages four independently trained models per encoder (two tasks by two budgets), and EquivDP3 exceeds PointNet+Aug in all four (margins 4.5--32 points). It is not the best encoder everywhere: DGCNN leads on Task 2 at $n_{demo}{=}5$ and PCT on Task 1 at $n_{demo}{=}10$.

This is why we frame our claim at the level of the low-data aggregate (14.7-point margin, Sec.~\ref{sec:experiments}-C) rather than any individual cell, and why we decline to claim the 3-point OOD-drop margin at $n_{demo}{=}100$. A properly seeded replication (three training seeds per cell with confidence intervals, prioritizing the $n_{demo}{\in}\{5,10\}$ cells that carry the argument) is the single most valuable addition to this study.

\subsection{Computational Cost}
\begin{figure}[t]
\centering
\includegraphics[width=\linewidth]{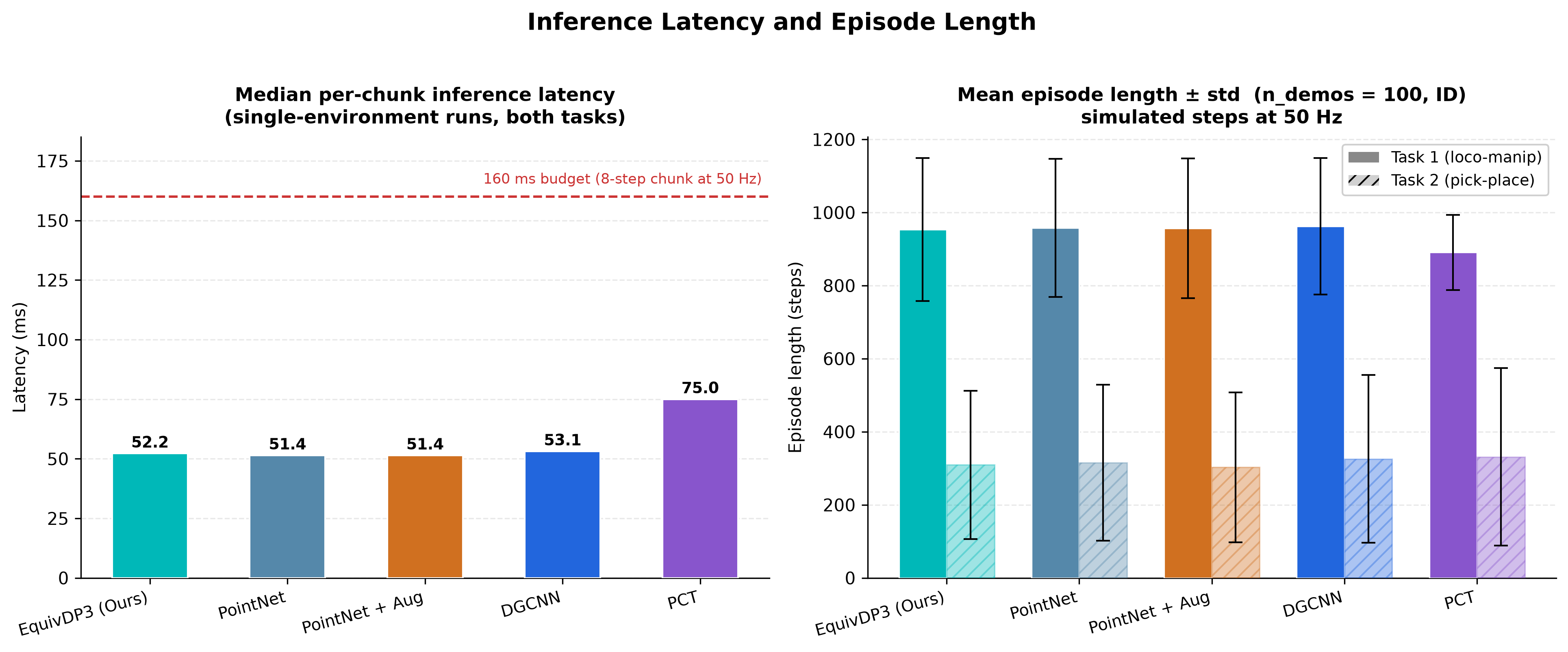}
\caption{Left: median per-chunk inference latency over all single-environment evaluation runs, well under the 160\,ms real-time budget (dashed line) implied by an 8-step chunk at 50\,Hz; EquivDP3 costs 0.8\,ms more than PointNet, and PCT is the slowest. Right: mean episode length ($\pm$std) in simulated steps at $n_{demo}{=}100$, ID, where every encoder succeeds on 92--100\% of episodes. Episode length largely restates success rate, so this panel is informative only at matched success, as here (Sec.~\ref{sec:experiments}-G).}
\label{fig:latency}
\end{figure}

A natural concern is whether the additional structure in the VNN encoder (vector-valued features, the extra normalization steps in Algorithm~\ref{alg:encoder}) imposes a runtime penalty relevant to real-time control. Fig.~\ref{fig:latency} reports median per-chunk inference latency over all single-environment evaluation runs, and mean episode length in simulated steps at $n_{demo}{=}100$, ID.

Median per-chunk inference latency is 52.2\,ms for EquivDP3, 51.4\,ms for PointNet and PointNet+Aug, 53.1\,ms for DGCNN, and 75.0\,ms for PCT; all five are within the 160\,ms real-time budget implied by re-planning every 8 low-level steps at 50\,Hz. These figures are measured on single-environment evaluation runs only. The restriction matters: the logged per-environment timing divides the batch cost by the number of parallel environments, and only the PCT sweep contains multi-environment runs, so comparing the per-environment figures directly makes PCT appear the cheapest encoder when on a like-for-like basis it is the most expensive.

Together, these results indicate that the SIM(3)-invariant encoder's data-efficiency benefits come at essentially no computational cost relative to the non-equivariant baselines: our encoder costs 0.8\,ms more per chunk than the PointNet encoder it replaces. We caution that this is an end-to-end measurement: the 10-step DDIM chain through a $[512,1024,2048]$ U-Net dominates the budget and masks any encoder-level difference, so this supports ``the encoder is negligible relative to the backbone,'' not ``all five encoders cost the same.'' Encoder parameter counts are 127k (ours), 76k (PointNet and PointNet+Aug), 117k (DGCNN), and 282k (PCT); our encoder is neither the largest nor the smallest, so its low-data advantage is not a capacity artifact.

We cannot, from the quantities we log, rule out a confound in which an encoder's apparent advantage comes from faster or more aggressive motion rather than better task completion. Episode length is not independent of success: successful episodes terminate on completion while failures run on, so mean episode length largely restates the success rate. DGCNN at $n_{demo}{=}10$ on Task 1 averages 1487 simulated steps at 4\% success, against 953 steps at 98\% success for EquivDP3 at $n_{demo}{=}100$. Testing the confound properly would need episode length conditioned on success, which we do not record. The one comparison that is meaningful is at matched success: at $n_{demo}{=}100$ on Task 1, where every encoder reaches 97--100\%, mean episode length spans only 891--963 simulated steps (EquivDP3: 953), so where the comparison can be made, no encoder wins by moving faster. We note also that failing episodes end before the 1800-step cap rather than timing out (back-solving mean episode length for the near-floor cells under a timeout assumption returns impossible negative per-success durations), indicating that failures terminate on a fall or abort condition.

\section{Discussion}
\label{sec:discussion}

\textbf{Why does the advantage concentrate at low data?} The intuitive explanation, that with enough demonstrations a non-equivariant encoder implicitly learns the relevant invariances and so erases the structural advantage, is appealing but, on Task 2, where we can test it, not what happens. Our proprioception-only control (Sec.~\ref{sec:experiments}-E) already reaches 92\% at $n_{demo}{=}50$ (EquivDP3: 90\%) and 94\% at $n_{demo}{=}100$ (EquivDP3: 96\%) \emph{without seeing the point cloud at all}, and under distribution shift it matches every encoder at both budgets. From $n_{demo}{=}50$ onward Task 2 is therefore solvable without vision, from proprioception and the demonstration prior alone; there, the convergence of all five encoders reflects a ceiling in the benchmark rather than five encoders independently discovering the same geometric structure. This matters for interpretation: the high-data cells measure how well a policy can replay a narrow demonstration distribution, not how well its encoder represents 3D geometry.

The low-data regime is where the perceptual comparison is actually informative, and there the same control confirms vision is required at both budgets: blind success is 31\% at $n_{demo}{=}5$ (29 points below EquivDP3, and above the weakest sighted encoder, PointNet at 10\%) and 78\% at $n_{demo}{=}10$, 21 points below EquivDP3's 99\%. So the gap our method opens at $n_{demo}{\in}\{5,10\}$ is a perceptual gap, not an artifact of a task that ignores its camera, even though that gap is already narrowing by $n_{demo}{=}10$. Why structure helps specifically there is the standard sample-complexity argument for invariant models (made rigorous for the linear case by Elesedy and Zaidi~\cite{elesedy2021provably}): a finite demonstration set spans few object poses, and a non-equivariant encoder must spend capacity and data learning pose-consistency that our encoder obtains by construction (Eqs.~\ref{eq:vnlinear}--\ref{eq:vnlrelu}).

\textbf{A caveat on the mechanism.} Our OOD condition randomizes the yaw of a single object within an otherwise fixed scene, plus lighting. This is \emph{not} a global SIM(3) transform of the input point cloud (the box moves relative to the table, bin, and robot), so the encoder's invariance guarantee does not formally cover it. Any OOD benefit we observe is therefore empirical rather than a consequence of the theory, and we do not claim otherwise. Testing a genuinely global transform (e.g.\ perturbing camera extrinsics, which \emph{is} covered by the guarantee) is the natural way to probe the structural claim directly, and we leave it to future work.

\textbf{A caveat on invariance.} Our encoder removes the cloud's global translation, rotation and scale, yet the planner must output wrist poses and base velocities in the robot frame, which depend on exactly those quantities. Pose information survives only through the relative arrangement of the object within the rest of the scene, which is also in the cloud. The architecture therefore has no principled reason to aid pose generalization, and the invariant latent may instead be acting as a low-variance scene code that makes a nominal trajectory easier to reproduce. Our results do not rule this out: the advantage is concentrated in distribution, and EquivDP3's ID-to-OOD drops at low data are among the largest we measure (32 and 33 points at $n_{demo}{=}5$ on Tasks 1 and 2). An equivariant action head, or an invariant latent paired with an equivariant pose read-out, is the natural test.

\textbf{Why does DGCNN sometimes collapse at very low data?} Table~\ref{tab:subgoals} narrows the question: at $n_{demo}{=}10$ on Task 1, DGCNN lifts the box and brings it within 0.6\,m of the bin on every episode yet completes the task on 4\%, so the failure lies in the final placement rather than in acquiring the object. We do not have a confirmed explanation. Raw capacity is an unlikely cause, since DGCNN has fewer encoder parameters than ours (117k vs.\ 127k), and the same encoder is the best on the simpler, fixed-base Task 2 at $n_{demo}{=}5$ (85\% ID, 66\% OOD), which suggests its difficulty is tied to the harder task rather than to the architecture being broadly unsuitable.

\textbf{Limitations.} We state these plainly, as several bear directly on how much weight the results can carry. \emph{(i) Statistical power.} No (task, budget) configuration is trained more than once; measured re-evaluation spread on identical checkpoints reaches 10 points (Sec.~\ref{sec:experiments}-F), so we confine our claim to the low-data aggregate and explicitly decline the individual-cell and OOD-drop comparisons that a single run would otherwise seem to license. \emph{(ii) Benchmark ceiling.} From $n_{demo}{=}50$ the Task 2 benchmark is solvable without vision in distribution, and under distribution shift a blind policy matches every encoder (Sec.~\ref{sec:blind}); those cells bound the benchmark, not the encoders. We have no blind control on Task 1. \emph{(iii) Theory--experiment gap.} Our OOD condition is a per-object yaw perturbation, not a global SIM(3) transform, so it lies outside what the encoder's invariance formally guarantees. \emph{(iv) Invariance, not equivariance.} Unlike EquiBot, our policy is not end-to-end equivariant: the encoder emits an invariant latent consumed by a conventional U-Net, so action prediction carries no structural guarantee, and the latent discards pose information the planner needs. \emph{(v) Scope.} Demonstrations are Mimic-generated from one (Task 1) or five (Task 2) human source demonstrations, so $n_{demo}$ counts generated episodes. All results are in simulation, with no sim-to-real evaluation on physical hardware and no tactile or force feedback for contact-rich grasps. \emph{(vi) Blind control replication.} Table~\ref{tab:blind} uses 100 episodes per $n_{demo}$, matching the sighted cells, across the full grid ($n_{demo}\in\{5,10,50,100\}$). The 29- and 21-point gaps at $n_{demo}{=}5,10$ are significant by Fisher's exact test; the differences at $n_{demo}{=}50$ and $100$ are within noise and should be read as ``at ceiling,'' not as exact rates. As elsewhere, each budget rests on a single training run.

\textbf{Future work.} The most immediate need is a properly seeded replication of the low-data cells (three training seeds with confidence intervals), which would convert our aggregate claim from suggestive to established. Second, the benchmark itself should be made perceptually demanding. Our OOD condition already widens the object's yaw, yet a blind policy still succeeds on roughly two episodes in three, so wider yaw alone is not enough; randomizing object position and scene layout until the proprioception-only control fails at \emph{all} demonstration budgets would make every cell perceptually meaningful. A blind control on Task 1 is the cheapest next measurement. Third, testing a global input transform (perturbed camera extrinsics) would probe the structural invariance claim directly, where the present OOD condition does not. Beyond that: physical deployment on a real Unitree G1, an end-to-end equivariant action head in the spirit of EquiBot or Equivariant Diffusion Policy~\cite{wang2024equidiff} rather than an invariant latent, and online RL fine-tuning of the planner~\cite{ren2024dppo,refinedp2026}.

\section{Conclusion}
\label{sec:conclusion}
We presented EquivDP3, a two-stage control architecture for humanoid loco-manipulation that carries EquiBot's SIM(3)-equivariant point-cloud encoder from wheeled mobile manipulators to the 43-joint Unitree G1, in the reduced form of an invariant perception latent conditioning a conventional diffusion backbone. The architecture separates a high-level diffusion planner from a low-level controller comprising a frozen reinforcement-learning locomotion policy and a model-based inverse-kinematics module for the arms.

Our central finding is a regime split. Averaged over two tasks and two distribution conditions, the invariant encoder leads the strongest baseline by 14.7 points at 5--10 demonstrations and trails it by 2.6 points at 50--100, and a proprioception-only control on Task 2 shows why: at the larger budgets that task is solvable without vision, both in distribution and under distribution shift, so those cells measure the task's ceiling rather than any encoder's representation. A subgoal decomposition places the low-data advantage in the manipulation phase, not in locomotion. Geometric structure in the perception backbone is therefore a worthwhile and nearly free lever when demonstrations are scarce, which is the regime that motivates imitation learning on humanoids in the first place. We report the negative half of this result alongside the positive one because the boundary between them is, in our view, the more useful contribution: it tells future work on this benchmark which cells are worth measuring.

\bibliographystyle{IEEEtran}
\bibliography{references}

\end{document}